\documentclass{article}
\usepackage{ijcai24}

\usepackage{times}
\usepackage{soul}
\usepackage{url}
\usepackage[hidelinks]{hyperref}
\usepackage[utf8]{inputenc}
\usepackage[small]{caption}
\usepackage{graphicx}
\usepackage{amsmath}
\usepackage{amsthm}
\usepackage{booktabs}
\usepackage{algorithm}
\usepackage[noend]{algorithmic}
\usepackage[switch]{lineno}
\usepackage{amsfonts}
\usepackage{tikz}
\usepackage{stmaryrd}
\usepackage{multirow}

\newtheorem{definition}{Definition}

\newcommand{\train}{$P_{\text{train}}$}
\newcommand{\test}{$P_{\text{test}}$}
\newcommand*\GOR{\ |\ }
\newcommand{\microrts}{MicroRTS}
\newcommand{\name}{\textsc{LISS}}

\title{Searching for Programmatic Policies in Semantic Spaces}

\author{
    Author Name
    \affiliations
    Affiliation
    \emails
    email@example.com
}

\author{
\author{
Rubens O. Moraes$^{1}$
\And
Levi H. S. Lelis$^2$$^,$$^3$
\affiliations
$^1$ Departamento de Inform\'atica, Universidade Federal de Vi\c{c}osa, Brazil\\
$^2$Department of Computing Science, University of Alberta, Canada \\
$^3$Alberta Machine Intelligence Institute (Amii)
\emails
rubens.moraes@ufv.br,
levi.lelis@ualberta.ca
}
}

\begin{document}

\maketitle

\begin{abstract}
Syntax-guided synthesis is commonly used to generate programs encoding policies. In this approach, the set of programs, that can be written in a domain-specific language defines the search space, and an algorithm searches within this space for programs that encode strong policies. In this paper, we propose an alternative method for synthesizing programmatic policies, where we search within an approximation of the language's semantic space. We hypothesized that searching in semantic spaces is more sample-efficient compared to syntax-based spaces. Our rationale is that the search is more efficient if the algorithm evaluates different agent behaviors as it searches through the space, a feature often missing in syntax-based spaces. This is because small changes in the syntax of a program often do not result in different agent behaviors. 
We define semantic spaces by learning a library of programs that present different agent behaviors. Then, we approximate the semantic space by defining a neighborhood function for local search algorithms, where we replace parts of the current candidate program with programs from the library. We evaluated our hypothesis in a real-time strategy game called \microrts. Empirical results support our hypothesis that searching in semantic spaces can be more sample-efficient than searching in syntax-based spaces. 
\end{abstract}

\section{Introduction}

Programmatic representations of policies for solving Markov Decision-Processes (MDPs) offer advantages over neural representations, such as the ability to better generalize to similar but different scenarios than those used in training~\cite{InalaBTS20}. Previous work has also argued that programmatic representations can allow for policies that are more amenable to interpretability~\cite{VermaMSKC18}. Nevertheless, programmatic representations pose a difficult hurdle, since programmatic policies are generated by searching in often very large and discontinuous spaces of programs. 

A commonly used approach to searching in the space of programs is local search algorithms~\cite{koza:book92,programsynthesis_sa,aleixo23,locallearner}. All programs that can be written in a domain-specific language form the space of candidate solutions for local search algorithms. The search starts in one of these candidate programs, and through a neighborhood function, the search decides which program to evaluate next. The search continues until reaching a local optimum or exhausting a search budget (e.g., the number of programs evaluated). Searching in the programmatic space is difficult not only because the space of programs is often vast, but also because the search lacks guidance. The neighborhood function is defined by making small modifications to the current candidate program, which often do not result in a change of behavior of the policy encoded in the resulting program. The search algorithm thus spends a considerable portion of its computational budget evaluating programs that are different, but that represent the same policy. 

In this paper, we present an alternative approach to defining the programmatic search space of local search algorithms. Instead of defining neighborhood functions where neighbors differ in syntax, we present a method to approximate the underlying semantic space of the language. That is, neighbors in the semantic space will differ in agent behavior instead of simply syntax. We hypothesized that local search algorithms searching in semantic spaces are more sample-efficient than the same algorithms searching in traditional syntax spaces. 

We consider the setting in which the agent learns programmatic policies for an MDP and transfers the knowledge learned to speed up learning in other MDPs. Specifically, our method learns a library of semantically different programs while generating a policy for the first MDP, which is then used to define approximations of semantic spaces for downstream MDPs. This approximation is achieved by defining a neighborhood function in which, instead of simply changing the current candidate program in terms of syntax, we replace parts of it with programs from our library of programs. 

We evaluated our hypothesis that local search algorithms are more sample-efficient in semantic spaces than in traditional spaces in the game of \microrts, a challenging real-time strategy game~\cite{ontanon2017combinatorial}. Our results show that neighbor programs in our library-induced space tend to be semantically different, while often neighbor programs in traditional spaces are semantically identical. The results also support our sample efficiency hypothesis, since Stochastic Hill Climbing (SHC) synthesized much stronger policies while searching in our semantic space than when searching in traditional spaces. Finally, the policies SHC synthesized while searching in our semantic spaces compared favorably with the winners of the last three \microrts\ competitions.\footnote{The implementation of our system is available online at \url{https://github.com/rubensolv/Library-Induced-Semantic-Spaces}} 

\section{Problem Definition}

The problems we solve can be represented as Markov decision processes (MDPs) $(S, A, p, r, \mu, \gamma)$, where $S$ represents the set of states and $A$ is the set of actions. The function $p(s_{t+1}|s_t, a_t)$ encodes the transition model since it gives the probability of reaching state $s_{t+1}$ given that the agent is in $s_t$ and takes action $a_t$ at time step $t$. The agent observes the reward value of $R_{t+1}$ when moving from $s_t$ to $s_{t+1}$; the function $r$ returns the reward after a state transition. $\mu$ represents the distribution of initial states of the MDP. Finally, $\gamma$ in $[0, 1]$ is the discount factor. A policy $\pi$ is a function that receives a state $s$ and an action $a$ and returns the probability in which $a$ should be taken in $s$. The goal is to learn a policy $\pi$ that maximizes the expected sum of discounted rewards for $\pi$ starting in $s_0$, an initial state sampled from $\mu$: $\mathbb{E}_{\pi,p,\mu}[\sum_{k=0}^\infty \gamma^k R_{k+1}]$. 

Let \train\ be an MDP, which we refer to as the training problem, for which the agent learns to maximize the expected sum of discounted rewards. After learning a policy for \train, we evaluate the agent while learning policies for another problem, \test. In this paper, we learn a semantic space while learning a policy for \train\ and use it to learn a policy for \test. 


\subsection{Programmatic Policies}

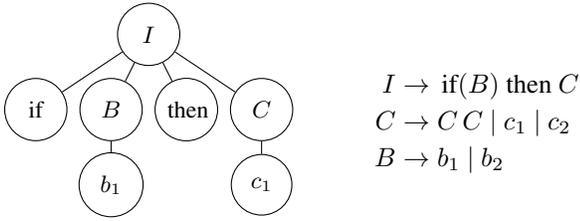
\begin{figure}[t]
\centering
\begin{minipage}{0.25\textwidth}
        \begin{tikzpicture}[level distance=10mm,sibling distance=10mm,every node/.style={text width=0.5cm,align=center,circle,draw,}]
        \small
            \node{\texttt{$I$}}
                child{node{if}}
                child{node{$B$}
                    child{node{$b_1$}}
                }
                child{node{then}}
                child{node{$C$}
                    child{node{$c_1$}}
                }
                ;
        \end{tikzpicture}
\end{minipage}
\begin{minipage}{0.2\textwidth}
\begin{align*}
I &\to \, \text{if}(B) \text{ then } C \, \\
C &\to C \, C \GOR c_1 \GOR c_2 \\
B &\to b_1 \GOR b_2 
\end{align*}
\end{minipage}
\caption{Abstract syntax tree for program ``if $b_1$ then $c_1$'' and the grammar defining the DSL.}
\label{fig:dsl}
\end{figure}

We consider programmatic policies, i.e., policies encoded in computer programs written in a domain-specific language (DSL). The programs a language accepts can be defined as a context-free grammar $(M, \Omega, R, I)$, where $M$, $\Omega$, $R$, and $S$ are the sets of non-terminals, terminals,
the production rules, and the grammar's initial symbol, respectively. Figure~\ref{fig:dsl} shows an example of a DSL (right), where $M = \{I, C, B\}$, with $I$ being the initial symbol, $\Omega = \{c_1, c_2, b_1, b_2$, if, then$\}$, $R$ are the production rules (e.g., $C \to CC$ and $B \to b_1$). 

Programs are represented as abstract syntax trees (AST). In such trees, each node $n$ and its children represent a production rule of the DSL if $n$ represents a non-terminal symbol. For example, the node $B$ and its child $b_1$, in Figure~\ref{fig:dsl}, represent the production rule $B \to b_1$. In the AST, every leaf node represents terminal symbols of the grammar. 
Figure~\ref{fig:dsl} shows an example of an AST for the program ``if $b_1$ then $c_1$''. A language $D$ defines the space of programs $\llbracket D \rrbracket$, which can be infinite. Often, in practice, the set of programs $\llbracket D \rrbracket$ can be defined to be finite. For example, one can define a maximum AST size (in terms of nodes in the tree) for the programs in $\llbracket D \rrbracket$. We consider DSLs such that the programs in $\llbracket D \rrbracket$ represent programmatic policies. The problem we tackle in this paper is to find a program encoding a policy that maximizes the expected sum of discounted rewards for an MDP. 

\section{Programmatic Search Spaces}
\label{sec:syntax}

In syntax-guided synthesis~\cite{sygus,pirl2018}, one searches in the space the DSL's grammar defines. 
%
A popular approach to syntax-guided synthesis, especially for solving reinforcement learning problems, is stochastic local search~\cite{pirl2018,VermaProjection2019,marino2021programmatic}. In stochastic local search, every program in $\llbracket D \rrbracket$ is a candidate solution (a program that attempts to maximize the sum of the rewards), and these candidates are related through a neighborhood function $\mathcal{N}_k(p)$ that receives a candidate $p$ and returns a set of $k$ neighbor candidates. The neighborhood function is often stochastic, which means that different calls to $\mathcal{N}_k(p)$ can return a different set of neighbors. Search algorithms, such as SHC, search in the space $D$ and $\mathcal{N}_k(p)$ induce for a program encoding a policy that maximizes the expected sum of rewards of an MDP. We define a search space for local search algorithms as follows. 

\begin{definition}[Search Space]
A search space is defined with a tuple $(D, \mathcal{N}_k, \mathcal{I}, \mathcal{E})$, where $D$ is a DSL with $\llbracket D \rrbracket$ defining the set of candidate solutions. $\mathcal{N}_k$ is a neighborhood function that receives a candidate in $\llbracket D \rrbracket$ and returns $k$ candidates from $\llbracket D \rrbracket$. $\mathcal{I}$ is a function that returns an initial candidate in $\llbracket D \rrbracket$. Finally, $\mathcal{E}$ is the evaluation function, that receives a candidate in $\llbracket D \rrbracket$ and returns a real value $\mathbb{R}$.
\end{definition}


A search space commonly used in the literature is defined as follows~\cite{koza:book92,marino2021programmatic}. The set of candidates is all programs in $\llbracket D \rrbracket$ whose AST has at least $z$ nodes, where $z$ is a hyperparameter. Controlling the size of the candidate programs offers an additional inductive bias. In our case, we prevent the generation of programs that are ``too simple''. The function $\mathcal{I}$ generates an initial candidate by starting with the initial symbol $I$ of the grammar and applying one of the production rules, chosen uniformly at random, to replace $I$. If the resulting string $p$ only contains terminal symbols, $\mathcal{I}$ returns $p$. Otherwise, $\mathcal{I}$ arbitrarily selects a non-terminal symbol $C$ in $p$ and replaces it by applying a production rule to $C$ that is also chosen uniformly at random from the available rules. This process is repeated while there is a non-terminal symbol in $p$. We enforce programs with at least $z$ nodes with a sampling rejection scheme. 

The neighborhood function $\mathcal{N}_k(p)$ is defined as follows. It selects, uniformly at random, a node $n$ in the AST of $p$ that represents a non-terminal symbol. The subtree rooted at $n$ is replaced in the AST by a subtree that is generated in a process similar to the one described for the function $\mathcal{I}$. We select uniformly at random a production rule that can be applied to the non-terminal symbol $n$ represents; this process is repeated to all non-terminals in the resulting string and it stops once the string only has terminal symbols. The process of replacing a subtree of $p$ is repeated $k$ times, thus generating $k$ neighbors. The evaluation function $\mathcal{E}$ is problem-dependent. For example, if the programmatic space encodes policies for solving MDPs, then $\mathcal{E}(p)$ approximates the expected sum of discounted rewards for program $p$, which can be approximated by rolling out $p$ from initial states sampled from $\mu$. We refer to this search space as the \textbf{syntax space} since it uses the syntax of the language to define the functions $\mathcal{N}_k$ and $\mathcal{I}$.



\subsection{Searching in Programmatic Search Spaces}

A popular approach to searching in programmatic spaces is with local search algorithms, such as SHC. For a given computational budget, SHC starts its search with the candidate $c$ that the function $\mathcal{I}$ returns; we refer to $c$ as the current candidate. In every iteration, it queries the neighborhood function $\mathcal{N}_k(c)$ and evaluates all neighbors of $c$ with respect to $\mathcal{E}$. Before starting a new iteration, SHC sets $c$ to be the neighbor with the best $\mathcal{E}$-value. The search stops and returns the current candidate $c$ if it reaches a local optimum, that is, none of the neighbors has a better $\mathcal{E}$-value than $c$. SHC is implemented with a restarting scheme: the search is restarted with a new candidate $\mathcal{I}$ returns whenever it reaches a local optimum and the algorithm has not exhausted its computational budget.   

The set of neighbors in the syntax space often allows neighbors to be syntactically similar, since they differ from the current candidate only by one subtree of their ASTs. This notion of syntactical locality is important because it can result in a sequence of locally improving programs that guide local search algorithms from the initial candidate to a candidate that maximizes the sum of discounted rewards for an MDP. 

\section{Semantic-Based Search Spaces}
\label{sec:semantic}

We hypothesize that local search algorithms will be more sample-efficient if the neighborhood function induces search spaces in which neighbors \emph{behave} differently, i.e., neighbor programs are semantically different. We base our hypothesis on the observation that algorithms searching in the syntax space spend a large portion of their computational budget evaluating programs that are syntactically different but are semantically identical. The evaluation of semantically identical programs can slow down the search for two reasons. First, the search uses its computational budget to evaluate the same behavior multiple times. Second, local search algorithms, such as SHC, are more effective when the evaluation function provides search guidance. For example, if all neighbors of a program $p$ are semantically identical to $p$, then the search is forced to restart, since all semantically identical programs evaluate to the same $\mathcal{E}$-value, thus representing a local optimum. The lack of search guidance can force the search to restart even if the search is near a promising program. We introduce the notion of $\beta$-proper spaces to measure the extent to which neighbors in a space are semantically different. 

\begin{definition}[$\beta$-proper]
Let $p_\beta$ be the probability that a program $p$, which is sampled uniformly at random from $\llbracket D \rrbracket$, has a neighbor that behaves identically to $p$. A search space is $\beta$-proper if its $p_\beta$-value is less than $\beta$, for a small $\beta$. 
\label{def:beta}
\end{definition}

The behavior of programs used in Definition~\ref{def:beta} depends on the application. For example, in sequential decision-making problems, behavior can be determined by the sequences of actions of two programs performed from a set of initial states.



\section{Semantic Spaces}

We introduce a method that uses a library of learned programs to induce $\beta$-proper spaces. Instead of generating neighbors by replacing a subtree of the current candidate program with another subtree randomly generated from the grammar of the DSL, we replace the subtree with a program from our library. The library is composed of semantically different programs, which are generated while searching for a policy that maximizes the reward in \train. We hypothesized that this library-induced space can be $\beta$-proper because the subtrees used to generate neighbors are themselves programs with well-defined behavior---subtrees for which we cannot evaluate their semantics (e.g., the code crashes) are not added to the library. When generating subtrees by randomly applying the rules of the grammar, as we do in the syntax space, we can generate subtrees that do not affect the program behavior. For example, the subtree ``if False then $c_1$'' does not change any program's behavior, independently of $c_1$. By using programs from the library, we guarantee that the subtree added to the program while generating one of its neighbors has a well-defined behavior. Moreover, the behavior of the program added is unique within the library of programs. 

\subsection{Library Construction}

\begin{algorithm}[t]
\caption{Library Construction}
\label{alg:library}
\begin{algorithmic}[1]
\REQUIRE Training problem \train, local search algorithm \textsc{LS}
\ENSURE A library $\mathcal{L}$ of semantically different programs.
\STATE $\mathcal{P}, \mathcal{S} \gets $\textsc{LS}$($\train$)$ \emph{\# using syntax space} \label{line:train}
\STATE $\mathcal{L} \gets \emptyset$ \label{line:init_l}
\STATE \textsc{Signature} $\gets \emptyset$ \label{line:init_signature}
\FOR{each $p$ in $\mathcal{P}$} \label{line:for_programs}
\FOR{each subtree $t$ rooted at a non-terminal in $p$} \label{line:for_subtrees}
\STATE $\mathcal{A}[s] \gets t(s)$ for each $s$ in $\mathcal{S}$ \label{line:compute_signature}
\IF{$\mathcal{A}$ is not in \textsc{Signature}} \label{line:verify_signature}
\STATE $\mathcal{L} \gets \mathcal{L} \cup \{t\}$ \label{line:grow_library}
\STATE \textsc{Signature} $\gets$ \textsc{Signature} $\cup \{\mathcal{A} \}$ \label{line:grow_signature}
\ENDIF
\ENDFOR
\ENDFOR
\RETURN $\mathcal{L}$
\end{algorithmic}
\end{algorithm}

Algorithm~\ref{alg:library} shows the process of constructing our library $\mathcal{L}$ of semantically different programs. The procedure receives the training problem \train\ and a local search algorithm \textsc{LS}, and it returns $\mathcal{L}$. In line~\ref{line:train}, the procedure invokes \textsc{LS} to search, using the syntax space, for a programmatic policy that maximizes the reward in \train. The search \textsc{LS} performs returns all programs encountered in the process, $\mathcal{P}$, and a set of states $\mathcal{S}$ encountered while evaluating programs in the \textsc{LS} search. In lines~\ref{line:init_l} and \ref{line:init_signature}, we initialize the library $\mathcal{L}$ and a set of action-signatures \textsc{Signatures}. An action-signature is a vector $\mathcal{A}$ with one entry for each $s$ in $\mathcal{S}$ containing the action $a$ a program returns for $s$. We use the action-signatures to approximate program semantics: if two programs have the same action-signature, we deem them as semantically identical. 

The procedure then iterates through all programs in $\mathcal{P}$ (line~\ref{line:for_programs}) and, for each $p$, it iterates through each subtree $t$ of the AST of $p$ that represents a non-terminal symbol (line~\ref{line:for_subtrees}). For example, the AST shown in Figure~\ref{fig:dsl} would have three subtrees: $I$, $B$, and $C$. We then evaluate the action $t$ returns for each $s$ in $S$ (line~\ref{line:compute_signature}), thus computing the action-signature of $t$. Note that if the program $t$ cannot be executed (e.g., the subtree does not represent a complete program), then we do not consider adding it to the library (this is not shown in the pseudocode). If the action-signature $\mathcal{A}$ of a program $t$ is not in \textsc{Signatures}, then we add $t$ to $\mathcal{L}$ and $\mathcal{A}$ to \textsc{Signatures}---$t$ is the program in $\mathcal{L}$ representing the behavior $\mathcal{A}$ defines. 

\subsection{Library-Induced Semantic Space}

We define a library-induced semantic space (\name) as a search space that is identical to the syntax space, except for its neighborhood function. Instead of generating neighbors by using the rules of the grammar, we use the programs in $\mathcal{L}$. Similarly to the neighborhood function of the syntax space, in \name\ we randomly select a node $n$ representing a non-terminal symbol $N$ in the AST of the current candidate. Then, we randomly select a program $t$ in $\mathcal{L}$ whose AST root also represents the non-terminal $N$; $t$ replaces the subtree rooted $n$ in the candidate program, thus generating a neighbor. All $k$ neighbors in the \name\ are generated with the same process. 

In practice, we do not rely entirely on \name, but on a mixture of \name\ and the syntax space. This is because, depending on \train\ and the policy generated for it, \name\ might not be able to reach a good portion of the original program space through its neighborhood function because some of the symbols might be missing from the library. During the search with \name, we generate neighbors using the function $\mathcal{N}_k$ of the syntax space with probably $\epsilon$ and, with probability $1 - \epsilon$, we generate them using the function $\mathcal{N}_k$ from \name. This guarantees that the search has access to all programs of the language, not only in the initialization of the search but also during the search, through the neighborhood function. 

The semantic space of \name\ continues to improve as we search for a solution to \test. We use the programs encountered while searching for a programmatic policy for \test\ to grow the library of programs. If the search encounters a program $p$ while searching for a solution to \test\ that is semantically different from all the programs in the library, then $p$ is added to the library, thus updating the semantic space used in the search. The idea is that the semantic space is continually learned, which could be helpful in settings where there is no split of training and testing problems, but just a stream of problems that the agent needs to learn how to solve. 

\section{Empirical Methodology}

In this section, we describe our methodology to evaluate the hypothesis that search algorithms operating in \name\ are more sample-efficient than when operating in the syntax space. 

\subsection{Problem Domain: \microrts}

We evaluate \name\ using the \microrts\ domain, a real-time strategy game designed for research. There is an active research community that uses \microrts\ as a benchmark to evaluate intelligent systems.\footnote{https://github.com/Farama-Foundation/MicroRTS/wiki} \microrts\ is a game played with real-time constraints and very large action and state spaces~\cite{lelis2021planning}. Each player controls two types of stationary units (Bases and Barracks) and four types of mobile units (Workers, Ranged, Light, and Heavy). Bases are used to store resources and train Workers. Barracks can train Ranged, Light, and Heavy units. Workers can build stationary units, harvest resources, and attack opponent units. Ranged, Light, and Heavy units have different amounts of hit points and inflict different amounts of damage to opponent units. Ranged units differ from each other by causing damage from far away. In \microrts, a match is played on a grid, which represents the map. Due to the different structures of the maps, different maps might require different policies to play the game.

Since \microrts\ is a two-player zero-sum game, one learns a policy through a self-play scheme. One of the simplest methods we can use is Iterated Best Response (IBR)~\cite{lanctot2017unified}. IBR starts with an arbitrary policy for one of the players and it approximates, by searching in the programmatic space, a best response to this initial policy. In the next iteration, IBR attempts to approximate a best response to the best response computed in the previous iteration. This process is repeated for a number of iterations, which is normally determined by a computational budget, and the last policy for each player is returned as the output of IBR. Self-play algorithms such as IBR require one to solve many MDPs. Every computation of a best response represents an MDP since the other player is fixed and can be seen as part of the environment. Instead of using IBR, we use Local Learner (2L), another self-play algorithm that was shown to synthesize strong programmatic policies for \microrts~\cite{locallearner}. 

We use six maps of different sizes in our experiments (the names of the maps reflect their names in the \microrts\ public repository): NoWhereToRun (NWR 9$\times$8), itsNotSafe (INS 15$\times$14), letMeOut (LMO 16$\times$8), Barricades (BRR 24$\times$24), Chambers (CHB 32$\times$32), and BloodBath.scmB (BBB 64$\times$64). The last is an adaption for \microrts\ of a map from the commercial game StarCraft. All these maps are used as \test\ in our experiments; we used the map basesWorkersA (24$\times$24) as \train. We consider two starting locations on each map. When evaluating two policies, to ensure fairness, each policy plays one match in both locations on the map.

The results are presented in terms of winning rate. The winning rate of a policy against a set of other policies is given by the number of victories added to half of the number of draws, divided by the total number of matches played, and multiplied by 100. For example, if a policy plays 10 matches, wins 2, draws 1, and loses 7, its winning rate is 25. We are interested in measuring the sample efficiency of the approaches in terms of the number of games played, so we will present plots showing the winning rate by the number of games. 

We use a domain-specific language developed for \microrts\ called Microlanguage~\cite{medeiros2022can}. This language includes high-level functions such as ``haverst'' and also loops that iterate through the players' units. The loops in the Microlanguage allow for implementing prioritization schema. This is because once an action is assigned to a unit, it cannot be replaced, so instructions appearing early in the loops will have higher priority than later instructions. 

\subsection{Experiments Performed}

We perform three experiments. The first experiment evaluates the values of $\beta$ for which the syntax and the semantic spaces are $\beta$-proper. We evaluate the ``pure'' version of \name, where we do not mix the neighborhood function of the semantic space with that of the syntax space. We generate $50$ programs by following the function $\mathcal{I}$ of the syntax space, and for each of these programs $p$, generate $1000$ neighbors $p'$ of $p$ according to the space's neighborhood function. Then, we roll-out $p$ and each of its $p'$s once in the following maps: NWR, LMO, and BRR. Since \microrts\ is deterministic, we can use the action-signature of $p$ and $p'$ in these roll-outs to approximate their behavior. If the action-signatures are identical, then we assume them to be semantically identical; they are not semantically identical otherwise. When evaluating $p$ and $p'$, we use another neighbor of $p$ as the other player in the roll-outs. This data approximates $p_\beta$ for both spaces. 

The second experiment evaluates our hypothesis that a local search algorithm searching in \name\ is more sample-efficient than the same algorithm searching in the syntax space. We use SHC with restarts as the algorithm in our experiments. We chose to use SHC because it was shown to perform well in the synthesis of programmatic policies for \microrts~\cite{locallearner}. We use $k= 1000$ in $\mathcal{N}_k$ and a limit of 400 seconds for SHC to return a best response; once it reaches this time limit, it returns the best policy it encountered across all restarts of the search. We induce \name\ with the library learned from all the programs generated in a single run of 2L with SHC on the basesWorkersA 24$\times$24 map. The action-signatures of the programs considered for the library were generated by evaluating them in 400 states. These 400 states are collected by randomly selecting pairs of policies representing best responses in the self-play process 2L executes and playing them on \train; every state encountered in these matches is added to the set, up to a total of 400 states. We use $\epsilon = 0.20$ for mixing the syntax and semantic spaces. We evaluated $\epsilon$ in $\{0.10, 0.20\}$ in preliminary experiments and found that $0.20$ performed better. We perform 30 independent runs of this experiment and present average winning rates and the 95\% confidence intervals. Each of the 30 independent runs uses a library generated with an independent run of 2L with SHC on the basesWorkerA 24$\times$24 map. 

As baselines for the second experiment, we use 2L with SHC in the syntax space (denoted 2L) and a version of 2L with SHC in the syntax space that initializes the search for best responses in each \test\ with the solution returned in \train, denoted 2L-I. The programmatic representation we use is known to generalize well across maps~\cite{aleixo23}, so 2L-I is expected to be a strong baseline. 

The third experiment compares the policies SHC searching in \name\ synthesizes with those of the winners of the last three \microrts\ competitions:\footnote{\url{https://sites.google.com/site/micrortsaicompetition}} COAC,\footnote{\url{https://github.com/Coac/coac-ai-microrts}} Mayari,\footnote{\url{https://github.com/barvazkrav/mayariBot}} and RAISocketAI.\footnote{https://github.com/sgoodfriend/rl-algo-impls} COAC and Mayari are programmatic policies human programmers wrote in Java for the \microrts\ competition. RAISocketAI is a Deep Reinforcement Learning (DRL) system trained with the Proximal Policy Optimization algorithm~\cite{schulman2017proximal}. We evaluate all these algorithms in all six \test\ maps. COAC, Mayari, and RAISocketAI were evaluated and trained only in the maps NWR and BBB. For the RAISocketAI agent, we used the models trained for maps of the same size for the remaining four maps. 

In the third experiment, we ran our system six times for each of these maps, with each run returning a programmatic policy. We then run a round-robin tournament among these six policies and select the one with the highest winning rate. This selected policy is the one we evaluate against the winners of the last three competitions by playing each of them five times at each of the two starting locations we consider, for a total of ten matches on each map. We repeat this entire process five times, and we present average results for the five runs. This experiment evaluates how well the policies generalize to unseen opponents by simulating a tournament. 

We used a dedicated number of computers with the following settings: 16 GB of RAM, i7-1165G7 CPUs at 2.80 GHz with 8 threads. We also use $z = 4$ in all our experiments.

\section{Empirical Results}

Next, we present the results of our three experiments. 

\subsection{Experiment 1: $p_\beta$}

The $p_\beta$-value for the syntax space was $0.19$ with a standard deviation of $0.15$, while the $p_\beta$-value for \name\ was $0.01$ with a standard deviation of $0.01$. These results support our hypothesis that our \name\ is $\beta$-proper, for $\beta$ as low as $0.02$. The value of $0.19$ for the syntax space is quite high, which means that almost 20\% of the neighbors are semantically identical to the current candidate program, forcing the search to evaluate equivalent policies multiple times, possibly slowing down the search. While the $p_\beta$-values provide indication that \name\ is more ``friendly'' to local search algorithms than the syntax space, one can imagine degenerate cases where the space is $\beta$-proper but the behaviors of neighboring candidates do not result in a space with good search guidance. Experiments 2 and 3 complement the results of Experiment 1 to evaluate how friendly \name\ is to local search. 

\subsection{Experiment 2: Sample Efficiency}

\begin{figure*}[ht]
    \centering
     \includegraphics[width=1.75\columnwidth]{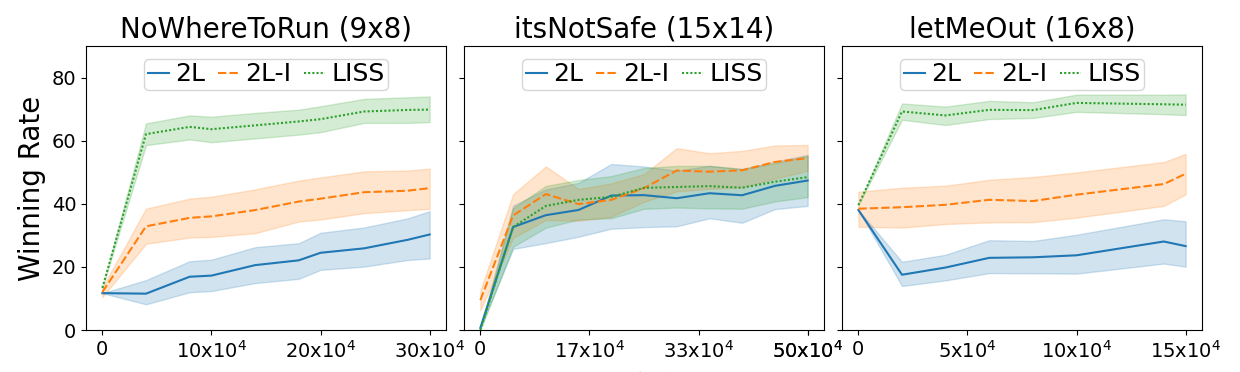}
     \includegraphics[width=1.75\columnwidth]{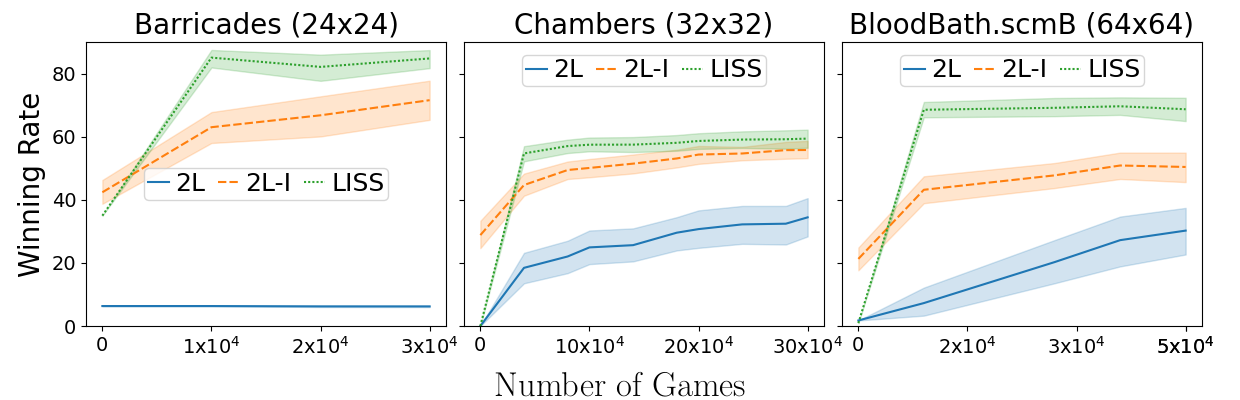}
    \caption{Results of each learning algorithm on six \microrts\ maps using 2L as the learning algorithm and Stochastic Hill Climbing as the search algorithm. These curves represent the winning rate of each algorithm compared to their opponents using the same amount of games.}
    \label{fig:microRTSSampleEficiency}
\end{figure*}

Figure~\ref{fig:microRTSSampleEficiency} presents the results. In these plots, we compute the winning rate of each method by having them play the last policy each of the other methods synthesizes. For example, on the Barricades map, the policy 2L synthesizes after playing $10000$ games is evaluated against the policies 2L-I and \name\ synthesize after playing $30000$ games, which is the maximum computational budget used in the experiments on this map. 

Before these experiments, 2L was the state-of-the-art synthesizer for policies for playing \microrts. The ability to start the search from a strong policy synthesized for \train\ already outperforms 2L on most maps by a large margin.  The use of the initialization of search with a strong policy for the baseWorkersA map makes a large difference in the Chambers map, where a policy similar to the one learned to play in \train\ is already quite strong. However, the semantic space of \name\ allows for a much more sample-efficient learning process, so \name\ quickly catches up and eventually synthesizes a policy that is stronger than that which 2L-I manages to synthesize. A similar pattern is observed on the largest map, but with a larger gap between 2L-I and \name. Overall, \name\ outperforms both baselines by a large margin on all but the itsNotSafe map, where there is no statistical difference between the methods. In this map, a simple policy, which trains a Ranged unit and attacks the opponent, is already very strong. This map essentially represents an easy search problem, that SHC is able to tackle searching in either the syntax space or \name. 

These results support our hypothesis that SHC is more sample-efficient while searching in \name\ than in the syntax space. Moreover, the results also suggest that searching in the semantic space of \name\ can be more effective than using a powerful initialization of the search in the syntax space. 

\begin{table*}[h]
\begin{tabular}{@{}lrrrrrrr@{}}
\toprule
\multirow{2}{*}{\textbf{Agents}} & \multicolumn{7}{c}{\textbf{Maps}}                                                                                                                                                     \\
\cmidrule{2-8}
                                 & \multicolumn{1}{l}{\textbf{NWR 9$\times$8}} & \multicolumn{1}{l}{\textbf{INS 15$\times$14}} & \multicolumn{1}{l}{\textbf{LMO 16$\times$8}} & \multicolumn{1}{l}{\textbf{BRR 24$\times$24}} & \multicolumn{1}{l}{\textbf{CHB 32$\times$32}} & \multicolumn{1}{l}{\textbf{BBB 64$\times$64}} & \multicolumn{1}{l}{\textbf{Avg. Agent}} \\
                                 \midrule
\textbf{COAC}                    & 55.00                            & 100.00                           & 72.50                           & 45.00                            & 37.50                            & 67.50                            & 62.92                                      \\
\textbf{Mayari}                  & 57.50                            & 95.00                            & 88.33                           & 65.00                            & 47.50                            & 90.00                            & 73.89                                      \\
\textbf{RAISocketAI}             & 40.83                            & 100.00                          & 60.00                           & 97.50                            & 100.00                           & 100.00                           & 83.06                                      \\
\midrule 
\textbf{Avg. Map}                 & 51.11                            & 98.33                            & 73.61                           & 69.17                            & 61.67                            & 85.83                           & 73.29                   \\
\bottomrule
\end{tabular}
\caption{Average winning rate of \name\ against the winner of the last three \microrts\ competitions (rows) in six maps (columns).}
\label{tbl:comp_results}
\end{table*}

\subsection{Experiment 3: Competition Agents}

Table~\ref{tbl:comp_results} shows the results of our third experiment. If we consider only the maps on which COAC, Mayari, and RAISocketAI were evaluated and trained (NWR and BBB), \name\ has a winning rate higher than 50 in all comparisons, but against RAISocketAI on the NWR map. NWR is a small map where DRL is able to learn a strong micromanagement of units (e.g., clever optimizations during combat). The ability to synthesize strong micromanagement of units is currently beyond the reach of \name\ due to a limitation of the Microlanguage that focuses on the macro aspects of the game (e.g., which combinations of units to train). \name\ obtains a winning rate of 100 against RAISocketAI on the larger BBB map. This highlights the difficulties of training DRL agents on larger problems. 

If we consider all six maps, then the comparisons in which \name\ has a winning rate smaller than 50 against a particular opponent are clearly outnumbered by the comparisons in which its rate is greater than 50. 
In addition to BBB, \name\ performs notably well on the INS map because the strong policies for this map do not generalize to other maps, despite being easy to find them, as discussed in the second experiment. However, COAC, Mayari, and RAISocketAI were not designed or trained for this map, which explains their performance. The policies SHC synthesizes by searching in the semantic space of \name\ outperform, on average, the winners of the last three competitions, thus demonstrating that these policies can generalize to unseen but similar MDPs, as the MDP effectively changes as we change the opponent.

\section{Related Works} 

Our research is related to previous work on Programmatically Interpretable Reinforcement Learning (PIRL)~\cite{VermaMSKC18}, where the objective is to generate human-readable programs encoding policies for addressing Reinforcement Learning tasks~\cite{BastaniPS18,VermaProjection2019}. Previous work in this area does not distinguish between syntactic and semantic spaces and performs the search for programs in spaces that are probably not $\beta$-proper for the small values of $\beta$ we observed in our experiments. Thus, the idea of learning a semantic space can potentially be applied to some of these works. In generalized planning one is interested in synthesizing programs encoding policies for solving classical planning problems~\cite{FSM-planning,SrivastavaIZZ11,Hu_DeGiacomo_2013,Aguas0J18}. Similarly to the work in PIRL, it would be interesting to investigate the idea of learning $\beta$-proper semantic spaces for solving classical planning problems with generalized planning. 

Previous work has also investigated the use of learned latent spaces for synthesizing programmatic policies~\cite{leaps2021,hprl2023}. In this line of work, a latent space is trained before the search starts by sampling programs from the grammar describing the language. The latent space is trained so that vectors near each other in the space represent programs that behave similarly. Like our work, this line of research is also concerned with the semantics of the programs. In contrast to our work, the latent space is not trained with the goal of attaining $\beta$-proper spaces. In fact, since the space is continuous, it is challenging to design neighborhood functions as one needs to choose the magnitude in which the vector representing the current candidate will be modified to generate a neighbor. Furthermore, recent work showed that the SHC we use in our experiments outperformed algorithms searching in learned latent spaces~\cite{carvalho2024reclaiming}.

Programmatic policies have also been used to guide tree search algorithms by reducing the action space in games such as \microrts. Puppet Search (PS)~\cite{barriga2017game} defines a search space, similar to the one defined by semantic space, by changing the parameter variables of handmade programmatic policies. Strategy Tactics (STT)~\cite{barriga2017combining} combines PS's search with a Na\"iveMCTS search~\cite{ontanon2017combinatorial} in a small fraction of the state space for combat micromanagement. 
Strategy Creation via Voting (SCV) generates policies via voting~\cite{SilvaMoraesLelis2018SCV}, which plays the game by combining also manually crafted policies. 
In contrast to PS, STT, and SCV which use manually crafted programmatic policies, \name\ introduces a method to synthesize programmatic policies. Another line of research uses programmatic policies for inducing action abstractions~\cite{Moraes2018}, where tree search algorithms consider only a subset of the actions available for search. This subset is determined by the actions a set of programmatic policies return at a given state. Future research might investigate the use of policies \name\ synthesizes to induce action abstractions for tree search algorithms. 

Dynamic Scripting (DS) is an algorithm for synthesizing programmatic policies for zero-sum role-playing games. DS generates programmatic policies by extracting rules from an expert-designed rule base according to a learned policy~\cite{spronck2004online}. DS has also been applied for zero-sum RTS games~\cite{ponsen2004improving,dahlbom2006goal}. Our method is more expressive because it considers spaces defined by DSLs, as opposed to being dependent on a particular DSL that allows one to define rules. 

DreamCoder and its extensions, Stitch and Babble, are systems that learn a library of programs in the context of supervised learning~\cite{dreamcoder,stitch,babble}. Our approach differs from these systems in important ways. We use a library of programs to define the search space of local search algorithms, while DreamCoder uses an enumerative approach guided by a learned function. Moreover, it is not clear how to learn a function to guide the search for programmatic policies. This is because we do not know the MDP we will solve ahead of time, before the agent interacts with the environment. By contrast, in DreamCoder's supervised learning setting, it is reasonable to assume that the training and testing problems come from similar distributions and an effective guiding function can be learned a priori. 

\section{Conclusions} 

We hypothesized that algorithms searching in programmatic spaces where neighbors encode similar but semantically different programs are more sample-efficient than algorithms searching in the syntax space. In this paper, we showed empirically that, often in syntax spaces, neighbors will be semantically identical, which could slow down the search. We then introduced Library-Induced Semantic Spaces (\name), where the neighbors of a candidate program in the space are generated by replacing parts of the candidate with programs from a library of semantically different programs. We showed empirically that \name\ is $\beta$-proper, for a small value of $\beta$, in the domain of \microrts, while the syntax space is not. Our results also supported our sample efficiency hypothesis, since the programs a local search algorithm synthesized by searching in \name\ encoded much stronger policies than the programs the same search algorithm synthesized while searching in the syntax space. We also showed empirically that the policies \name\ synthesizes can outperform the winners of the last three \microrts\ competitions, which include two programmatic policies written by human programmers and a DRL agent. Overall, our results suggest that the design of semantic spaces is a promising direction for methods that rely on search algorithms for synthesizing programmatic policies.

\section*{Acknowledgments}

This research was supported by Canada's NSERC, and the CIFAR AI Chairs program, and Brazil's CAPES. The research was carried out using computational resources from the Digital Research Alliance of Canada and the UFV Cluster. We thank the anonymous reviewers for their feedback.

\bibliographystyle{named}
\bibliography{ijcai24}

\end{document}


\maketitle

\section{Domain-Specific Language (DSL)}

In this section, we present the DSL used in our experiments. 

\subsection{\microrts}

The DSL we use for MicroRTS was introduced by \cite{marino2021programmatic} and expanded by \cite{medeiros2022can}. This DSL accepts nested if structures and nested for loops. The DSL is described with the following context-free grammar:
\begin{align*}
S &\to SS \GOR \text{for S} \GOR \text{if}(B) \text{ then } S \GOR \text{if}(B) \text{ then } S \text{ else } S\\
&\to C \GOR \lambda \\
B &\to b_1(T, N) \GOR b_2(T, N)  \GOR b_3(T, N) \GOR b_4(N) \GOR b_5(N)  \\
&\to \GOR b_6(N) \GOR b_7(T) \GOR b_8 \GOR b_9 \GOR b_{10} \GOR b_{11} \GOR b_{12} \GOR b_{13} \GOR b_{14} \\
C &\to c_1(T, D, N) \GOR c_2(T, D, N)  \GOR c_3(T_p, O_p) \GOR c_4(O_p)  \\
&\to c_5(N) \GOR c_6 \GOR c_7 \\
T &\to \text{Base} \GOR \text{Barracks} \GOR \text{Ranged} \GOR \text{Heavy} \GOR \text{Light} \\
&\to \text{Worker} \\
N &\to 0 \GOR 1 \GOR 2 \GOR 3 \GOR 4 \GOR 5 \GOR 6 \GOR 7 \GOR 8 \GOR 9 \GOR 10 \GOR 15 \GOR 20 \GOR 25 \\
&\to 50 \GOR 100 \\
D &\to \text{EnemyDir} \GOR \text{Up} \GOR \text{Down} \GOR \text{Right} \GOR \text{Left} \\ 
O_p &\to \text{Strongest} \GOR \text{Weakest} \GOR \text{Closest} \GOR \text{Farthest}  \\
&\to \text{LessHealthy} \GOR \text{MostHealthy} \GOR \text{Random} \\
T_p &\to \text{Ally} \GOR \text{Enemy} \\ 
\end{align*}

This DSL contains several Boolean functions (B) and command-oriented functions (C) that provide either information about the current state of the game or commands for the units the player controls. Here is the list of functions in the DSL. 

\begin{itemize}
\item\emph{$b_1(T,N)$}: Checks if the ally player has $N$ units of type $T$ (HasNumberOfUnits).
\item\emph{$b_2(T,N)$}: Checks if the opponent player has $N$ units of type $T$ (OpponentHasNumberOfUnits).
\item\emph{$b_3(T,N)$}:  Checks if the ally player has less than $N$ units of type $T$ (HasLessNumberOfUnits).
\item\emph{$b_4(N)$}: Checks if the ally player has $N$ units attacking the opponent (HaveQtdUnitsAttacking).
\item\emph{$b_5(N)$}: Checks if the ally player has a unit within a distance $N$ from an opponent's unit (HasUnitWithinDistanceFromOpponent).
\item\emph{$b_6(N)$}: Checks if the ally player has $N$ units of type Worker harvesting resources (HasNumberOfWorkersHarvesting).
\item\emph{$b_7(T)$}: Checks if a unit is an instance of  Type $T$ (is\_Type). 
\item\emph{$b_8$}: Checks if a unit is of type Worker (IsBuilder).
\item\emph{$b_9$}: Checks if a unit can attack (CanAttack).
\item\emph{$b_{10}$}: Checks if the ally player has a unit that kills an opponent's unit with one attack action (HasUnitThatKillsInOneAttack).
\item\emph{$b_{11}$}: Checks if the opponent player has a unit that kills an ally's unit with one attack action (OpponentHasUnitThatKillsUnitInOneAttack).
\item\emph{$b_{12}$}: Checks if an unit of the allied player is within attack range of an opponent's unit (HasUnitInOpponentRange).
\item\emph{$b_{13}$}: Checks if an unit of the opponent player is within attack range of an ally's unit (OpponentHasUnitInPlayerRange).
\item\emph{$b_{14}$}: Checks if a unit can harvest resources (CanHarvest).
\end{itemize}

Next, we describe the command-oriented functions used in our DSL:

\begin{itemize}
\item\emph{$c_1(T,D,N)$}: Trains $N$ units of type $T$ on a cell located on the $D$ direction of the unit (Build).
\item\emph{$c_2(T,D,N)$}: Trains $N$ units of type $T$ on a cell located on the $D$ direction of the structure responsible for training them (Train).
\item\emph{$c_3(T_p,O_p)$}:  Commands a unit to move towards the player $T_p$ following a criterion $O_p$ (moveToUnit).
\item\emph{$c_4(O_p)$}: Commands a unit to attack units of the opponent player following a criterion $O_p$ (Attack).
\item\emph{$c_5(N)$}: Sends $N$ Worker units to harvest resources (Harvest).
\item\emph{$c_6$}: Commands a unit to stay idle and attack if an opponent unit comes within its attack range (Idle).
\item\emph{$c_7$}: Commands a unit to move in the opposite direction of the player's base (MoveAway).
\end{itemize}

$T$ represents the type of a unit. $N$ is a set of integers. $D$ represents directions required as parameters for some of the functions. $O_p$ is a set of criteria for selecting an opponent unit based on the current state of the unit. $T_p$ represents the set of players. See the DSL above for all types, integers, directions, criteria, and players the DSL considers.

\section{MicroRTS Maps}

\begin{figure}[ht]
    \centering
    \includegraphics[width=.20\textwidth]{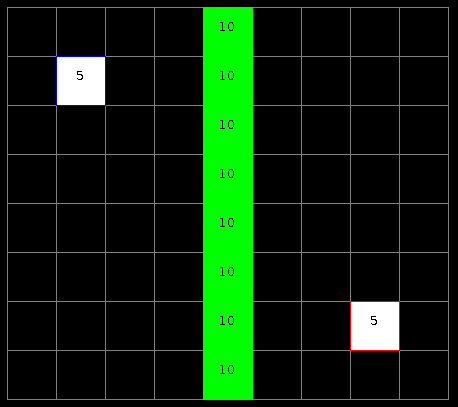}    
    \includegraphics[width=.20\textwidth]{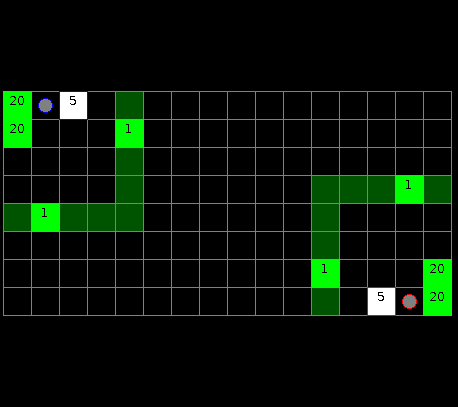}
    \includegraphics[width=.22\textwidth]{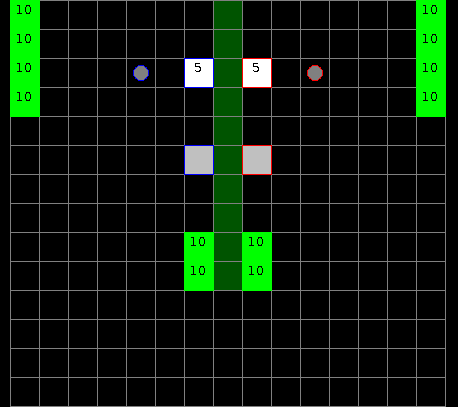}
    \includegraphics[width=.20\textwidth]{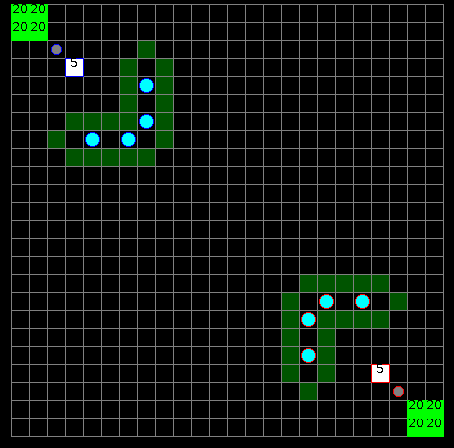}
    \includegraphics[width=.20\textwidth]{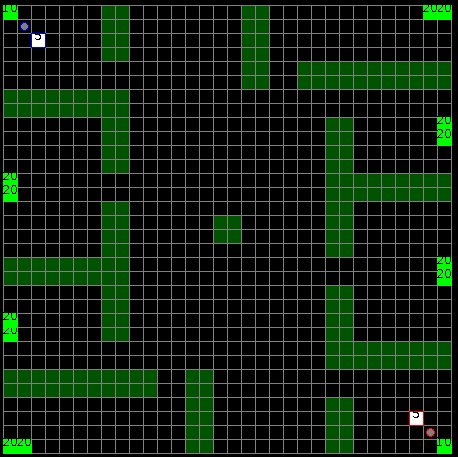}
    \includegraphics[width=.20\textwidth]{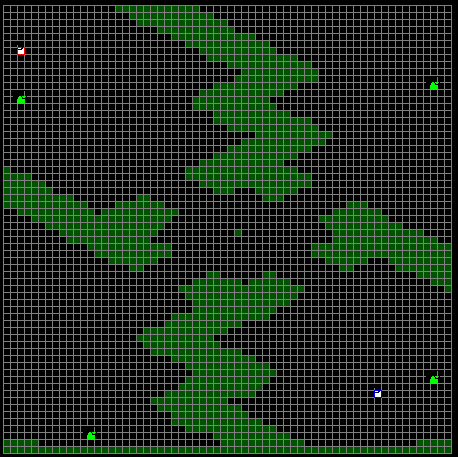}
    \caption{MicroRTS maps from 9$\times$8 (left, top) to 64$\times$64 (right, bottom).}
    \label{fig:MicroRTSMapsSupplement}
\end{figure}

Figure \ref{fig:MicroRTSMapsSupplement} shows the six maps used in our experiments (the names of the maps reflect their names in the \microrts\ public repository): NoWhereToRun (NWR 9$\times$8), itsNotSafe (INS 15$\times$14), letMeOut (LMO 16$\times$8), Barricades (BRR 24$\times$24), Chambers (CHB 32$\times$32), and BloodBath.scmB (BBB 64$\times$64).

\section{Detailed Tournament Evaluation}

Table~\ref{tbl:compe_results} presents the results of \name\ against the last three winners of the MicroRTS competition, Mayari, COAC, and RAISocketAI, for each map used in our simulated competition. Each map is ordered according to the average winning rate (Avg.) of each system. The average values shown in Table~\ref{tbl:compe_results} are slightly different from those shown in the paper because here they also account for the winning rate of $50.00$ of each system playing with itself. 

\begin{table}[!ht]
\setlength{\tabcolsep}{2.8pt}
\begin{tabular}{@{}lrrrrr@{}}
\toprule
\multicolumn{6}{c}{NoWhereToRun (NWR 9$\times$8)}\\ 
\midrule
 \textbf{Method} & \multicolumn{1}{l}{\textbf{COAC}} & \multicolumn{1}{l}{\textbf{RAISocketAI
}} & \multicolumn{1}{l}{\textbf{Mayari
}} & \multicolumn{1}{l}{\textbf{\name
}} & \multicolumn{1}{l}{\textbf{Avg.}} \\ 
Mayari  & 75,00  & 0,00  & 50,00  & 42,50  & 41,88 \\
COAC  & 50,00  & 50,00  & 25,00  & 45,00  & 42,50 \\
\name  & 55,00  & 40,83  & 57,50  & 50,00  & 50,83 \\
RAISocketAI  & 100,00  & 50,00  & 100,00  & 59,17  & 77,29 \\
\end{tabular}
\begin{tabular}{@{}lrrrrr@{}}
\toprule
\multicolumn{6}{c}{itsNotSafe (INS 15$\times$14)}\\ 
\midrule
 \textbf{Method} & \multicolumn{1}{l}{\textbf{COAC}} & \multicolumn{1}{l}{\textbf{RAISocketAI
}} & \multicolumn{1}{l}{\textbf{Mayari
}} & \multicolumn{1}{l}{\textbf{\name
}} & \multicolumn{1}{l}{\textbf{Avg.}} \\ 
CoaC  & 50,00  & 50,00  & 25,00  & 0,00  & 31,25 \\
Mayari  & 75,00  & 0,00  & 50,00  & 5,00  & 32,50 \\
RAISocketAI  & 100,00  & 50,00  & 100,00  & 0,00  & 62,50 \\
\name  & 100,00  & 100,00  & 95,00  & 50,00  & 86,25 \\
\end{tabular}
\begin{tabular}{@{}lrrrrr@{}}
\toprule
\multicolumn{6}{c}{letMeOut (LMO 16$\times$8)}\\ 
\midrule
 \textbf{Method} & \multicolumn{1}{l}{\textbf{COAC}} & \multicolumn{1}{l}{\textbf{RAISocketAI
}} & \multicolumn{1}{l}{\textbf{Mayari
}} & \multicolumn{1}{l}{\textbf{\name
}} & \multicolumn{1}{l}{\textbf{Avg.}} \\ 
CoaC  & 50,00  & 50,00  & 25,00  & 27,50  & 38,13 \\
Mayari  & 75,00  & 0,00  & 50,00  & 40,00  & 41,25 \\
RAISocketAI  & 100,00  & 50,00  & 100,00  & 11,67  & 65,42 \\
\name  & 72,50  & 88,33  & 60,00  & 50,00  & 67,71 \\
\end{tabular}
\begin{tabular}{@{}lrrrrr@{}}
\toprule
\multicolumn{6}{c}{ Barricades (BRR 24$\times$24)}\\ 
\midrule
 \textbf{Method} & \multicolumn{1}{l}{\textbf{COAC}} & \multicolumn{1}{l}{\textbf{RAISocketAI
}} & \multicolumn{1}{l}{\textbf{Mayari
}} & \multicolumn{1}{l}{\textbf{\name
}} & \multicolumn{1}{l}{\textbf{Avg.}} \\ 
Mayari  & 75,00  & 0,00  & 50,00  & 35,00  & 40,00 \\
CoaC  & 50,00  & 50,00  & 25,00  & 55,00  & 45,00 \\
RAISocketAI  & 100,00  & 50,00  & 100,00  & 2,50  & 63,13 \\
\name  & 45,00  & 97,50  & 65,00  & 50,00  & 64,38 \\
\end{tabular}
\begin{tabular}{@{}lrrrrr@{}}
\toprule
\multicolumn{6}{c}{Chambers (CHB 32$\times$32)}\\ 
\midrule
 \textbf{Method} & \multicolumn{1}{l}{\textbf{COAC}} & \multicolumn{1}{l}{\textbf{RAISocketAI
}} & \multicolumn{1}{l}{\textbf{Mayari
}} & \multicolumn{1}{l}{\textbf{\name
}} & \multicolumn{1}{l}{\textbf{Avg.}} \\ 
Mayari  & 75,00  & 0,00  & 50,00  & 52,50  & 44,38 \\
CoaC  & 50,00  & 50,00  & 25,00  & 62,50  & 46,88 \\
\name  & 37,50  & 100,00  & 47,50  & 50,00  & 58,75 \\
RAISocketAI  & 100,00  & 50,00  & 100,00  & 0,00  & 62,50 \\
\end{tabular}
\begin{tabular}{@{}lrrrrr@{}}
\toprule
\multicolumn{6}{c}{BloodBath.scmB (BBB 64$\times$64)}\\ 
\midrule
 \textbf{Method} & \multicolumn{1}{l}{\textbf{COAC}} & \multicolumn{1}{l}{\textbf{RAISocketAI
}} & \multicolumn{1}{l}{\textbf{Mayari
}} & \multicolumn{1}{l}{\textbf{\name
}} & \multicolumn{1}{l}{\textbf{Avg.}} \\ 
Mayari  & 75,00  & 0,00  & 50,00  & 10,00  & 33,75  \\
CoaC  & 50,00  & 50,00  & 25,00  & 32,50  & 39,38  \\
RAISocketAI  & 100,00  & 50,00  & 100,00  & 0,00  & 62,50  \\
\name  & 67,50  & 100,00  & 90,00  & 50,00  & 76,88  \\
\bottomrule
\end{tabular}
\caption{Average winning rate of the row agent against the column agents. \name\ is evaluated with the winners of the last three \microrts\ competitions.}
\label{tbl:compe_results}
\end{table}

\section{Synthesized Programs for MicroRTS}
Tables \ref{tab:LISSMap64}, \ref{tab:2LiMap64}, and \ref{tab:2LMap64} show high-performing programmatic policies each method synthesized (\name, 2L-I, and 2L) for the map BBB. The policies presented in the tables below are the ones with the highest winning rate in all 30 independent runs of each system. For example, the policy shown in Table \ref{tab:LISSMap64} trains four workers and sends them to collect resources as soon as they are trained.
 Once they have collected ``enough resources,'' they build one barrack. Once the barrack is ready, it starts to train Heavy units. The policy of \name\ is more refined than the one produced by 2L-I, which is shown in Table \ref{tab:2LiMap64}. \name's policy is optimized to produce only one barrack (a costly structure in the game), while 2L-I's policy produces up to seven barracks, if given enough resources. \name's policy uses \texttt{MoveToUnit} commands according to conditionals to define a refined control for the combat units, an aspect not found in the 2L-I policy. 

 \begin{figure}[ht]
    \centering
    \includegraphics[width=0.43\textwidth]{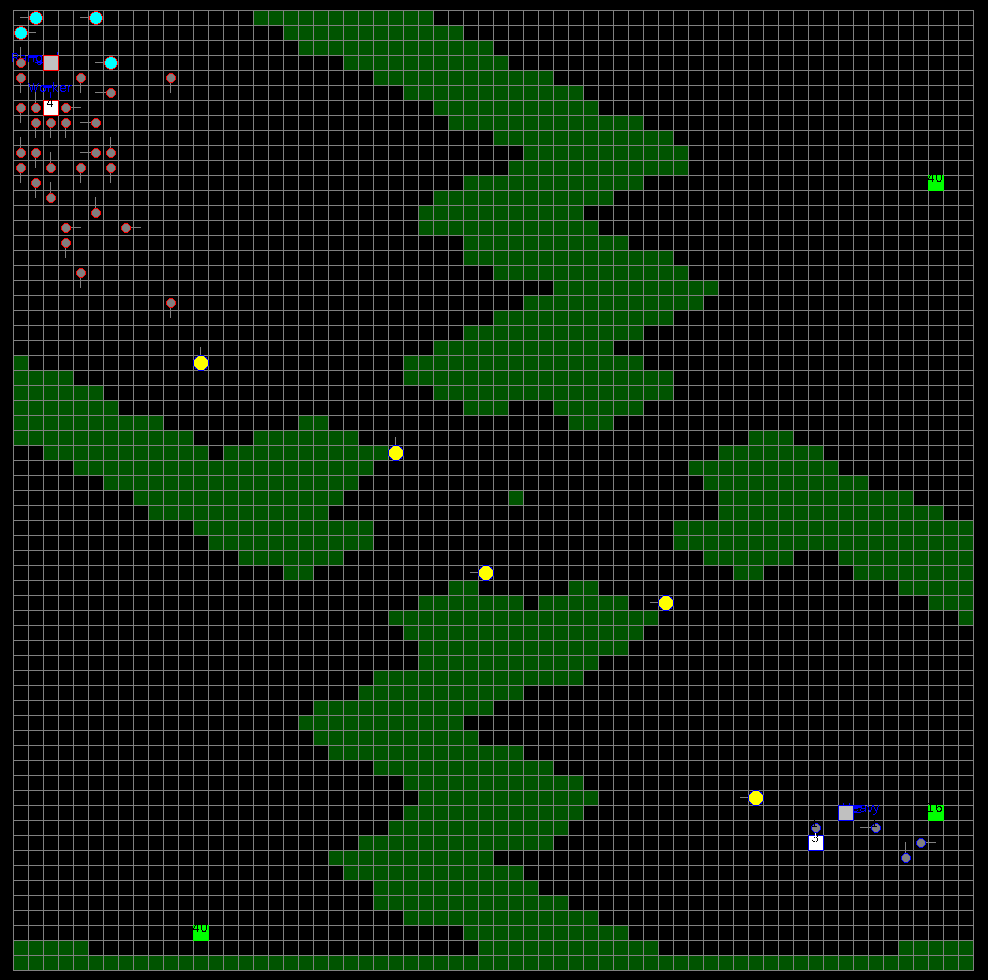}    
    \includegraphics[width=0.43\textwidth]{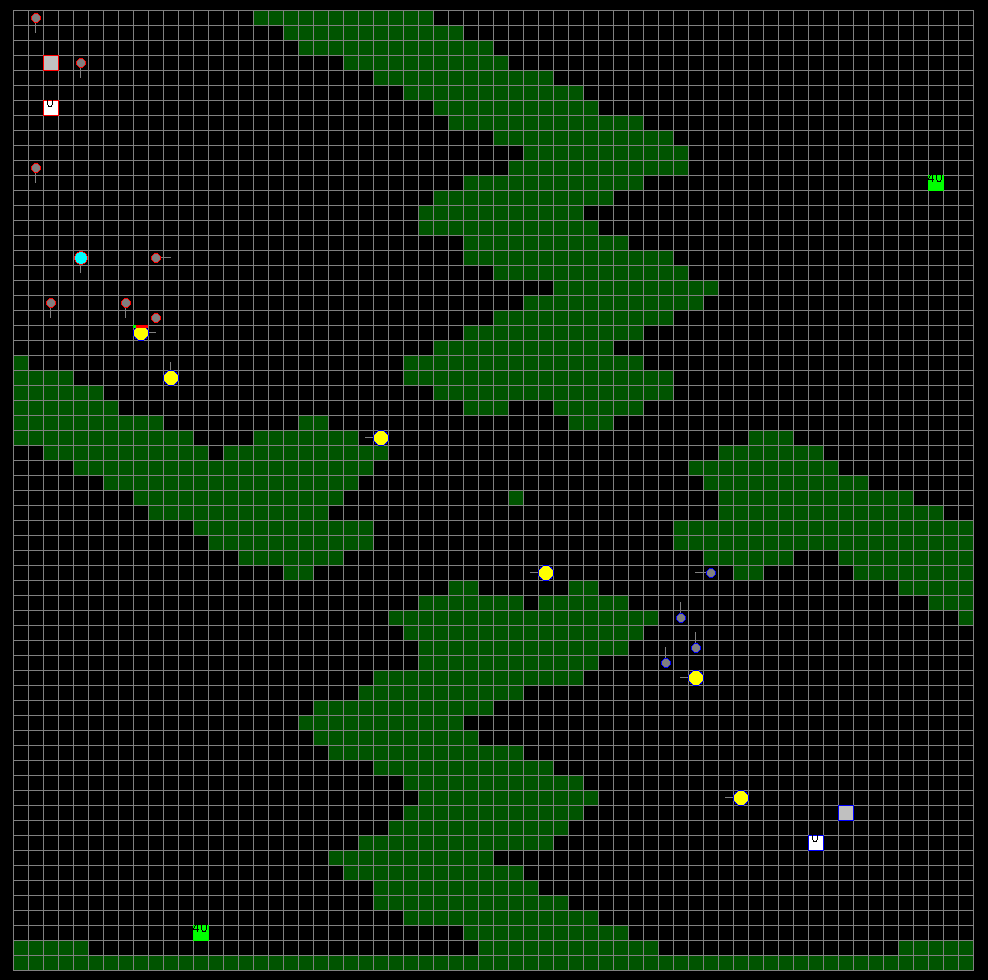}
    \caption{Match between a policy of \name\ (blue; bottom-right corner) against RAISocketAI (red; top-left corner). }
    \label{fig:LissVsRAISocketAI}
\end{figure}
 
Figure \ref{fig:LissVsRAISocketAI} shows two
game states of a match between \name's policy (bottom-right corner), against the RAISocketAI agent (top-left corner). In the top image, we can see how \name\ optimizes the training of Heavy units (larger yellow circles) using a few workers (small gray circles), saving resources for training a large number of Heavy units. This is in contrast to RAISocketAI's policy, which trains a large number of worker units, thus leaving fewer resources for combat units. In the bottom image, we can see a winning position of \name, since it controls a good number of Heavy units while RAISocketAI only has one Ranged unit (light blue circle) and many weaker Worker units. 
\name's policy was able to defeat RAISocketAI's on this map for all independent runs of the system.

\begin{table}[t!]
\begin{lstlisting}
def LISS-Solution-BloodBath-scmB-64$\times$64()
	for(Unit u)
		for(Unit u)
			u.attack(Closest)
			for(Unit u)
				u.harvest(1)
			for(Unit u)
				u.build(Barracks,EnemyDir,1)
			for(Unit u)
				u.harvest(9)
			for(Unit u)
				u.idle()			
			u.train(Heavy,Right,50)
			u.moveToUnit(Enemy,LessHealthy)
			if(u.HasUnitWithinDistFromOp(25)):
			     empty
		  else
				for(Unit u)
					u.attack(Weakest)
			if(u.OpHasUnitKillsInOneAttack()):
				u.moveToUnit(Ally,Farthest)
			u.moveToUnit(Enemy,Farthest)
			if(HasLessNumberOfUnits(Barracks,25)): 
				for(Unit u)
					u.train(Worker,EnemyDir,4)
					for(Unit u)
						u.attack(Closest)
						u.moveToUnit(Ally,Weakest)
			else 
				u.moveToUnit(Ally,MostHealthy)
			for(Unit u)
				u.moveToUnit(Ally,Closest)
				u.moveToUnit(Ally,Farthest)
		if(u.HasUnitWithinDistFromOp(15)):
		      empty
	  else
		      u.moveToUnit(Enemy,LessHealthy)
	
\end{lstlisting}
\caption{Program \name\ synthesized for map BloodBath.scmB (BBB 64$\times$64).}
\label{tab:LISSMap64}
\end{table}

\begin{table}[t!]
\begin{lstlisting}
def 2L-i-Solution-BloodBath-scmB-64$\times$64()
	for(Unit u)
		u.train(Heavy,Up,4)
		u.build(Barracks,Down,7)
		u.idle()
		u.harvest(2)
	for(Unit u)
		for(Unit u)
			u.train(Worker,Down,5)
			u.harvest(5)
			u.attack(Strongest)
		for(Unit u)
			u.train(Heavy,Right,10)
 
\end{lstlisting}
\caption{Program 2L-i synthesized for map BloodBath.scmB (BBB 64$\times$64).}
\label{tab:2LiMap64}
\end{table}

\begin{table}[t!]
\begin{lstlisting}
def 2L-Solution-BloodBath-scmB-64$\times$64():
  for(Unit u)
		u.train(Worker,Left,20)	
	for(Unit u)
		u.idle()
	for(Unit u)
		u.harvest(6)
	for(Unit u){
		if(u.HasUnitWithinDistFromOp(20)):
			u.attack(MostHealthy)
		else
			u.attack(LessHealthy)
\end{lstlisting}
\caption{Program 2L synthesized for map BloodBath.scmB (BBB 64$\times$64).}
\label{tab:2LMap64}
\end{table}

\bibliographystyle{named}
\bibliography{ijcai24}